\documentclass{article}
\usepackage{spconf,amsmath,epsfig, cite}
\usepackage[utf8]{inputenc} 
\usepackage[T1]{fontenc}    
\usepackage{hyperref}       
\usepackage{url}            
\usepackage{booktabs}       
\usepackage{amsfonts}       
\usepackage{nicefrac}       
\usepackage{microtype}      
\usepackage{tcolorbox}
\usepackage{xcolor} 

\usepackage[marginal]{footmisc}
\usepackage{algorithmicx}
\usepackage{algorithm}
\usepackage{amsthm}
\usepackage{algpseudocode}
\usepackage{amsmath} 
\usepackage{graphicx}
\usepackage{subcaption}
\usepackage{multirow}
\usepackage{wrapfig}
\usepackage{pifont}       
\usepackage{bbding}       
\usepackage{fontawesome}  
\usepackage{color, colortbl}
\definecolor{LightGray}{rgb}{0.92,0.92,0.92}
\definecolor{Red}{rgb}{1.0, 0.13, 0.32}

\newcommand{\gree}[1]{{\color[rgb]{0,0,1}{\tiny\textbf{ZY:}}{\normalsize\itshape#1}}}
\newcommand{\eat}[1]{}

\newcommand{\E}{\mathbb{E}}

\newcommand{\dv}{{\boldsymbol d}}

\newcommand{\gv}{{\boldsymbol g}}

\newcommand{\pv}{{\boldsymbol p}}

\newcommand{\xv}{{\boldsymbol x}}
\newcommand{\yv}{{\boldsymbol y}}

\newcommand{\zv}{{\boldsymbol z}}

\newcommand{\deltav}{{\boldsymbol \delta}}
\newcommand{\thetav}{{\boldsymbol \theta}}
\newcommand{\lambdav}{{\boldsymbol \lambda}}

\newcommand{\Lcal}{\mathcal{L}}

\newcommand{\Dcal}{\mathcal{D}}

\let\OLDthebibliography\thebibliography
\renewcommand\thebibliography[1]{
  \OLDthebibliography{#1}
  \setlength{\parskip}{0pt}
  \setlength{\itemsep}{0pt plus 0.3ex}
}

\begin{document}\sloppy

\def\x{{\mathbf x}}
\def\L{{\cal L}}

\title{Adversarial Training with OCR modality Perturbation for Scene-Text Visual Question Answering}
%
\name{Zhixuan Shen, Haonan Luo$^{\ast}$\thanks{$^{\ast}$Corresponding author}, Sijia Li, Tianrui Li}
\address{School of Computing and Artificial Intelligence, Southwest Jiaotong University \\
shenzx29@gmail.com, lhn@swjtu.edu.cn, sijiali.centralemarseille@gmail.com, trli@swjtu.edu.cn
}

\maketitle

\begin{abstract}
Scene-Text Visual Question Answering (ST-VQA) aims to understand scene text in images and answer questions related to the text content. Most existing methods heavily rely on the accuracy of Optical Character Recognition (OCR) systems, and aggressive fine-tuning based on limited spatial location information and erroneous OCR text information often leads to inevitable overfitting. In this paper, we propose a multimodal adversarial training architecture with spatial awareness capabilities. Specifically, we introduce an Adversarial OCR Enhancement (AOE) module, which leverages adversarial training in the embedding space of OCR modality to enhance fault-tolerant representation of OCR texts, thereby reducing noise caused by OCR errors. Simultaneously, We add a Spatial-Aware Self-Attention (SASA) mechanism to help the model better capture the spatial relationships among OCR tokens. Various experiments demonstrate that our method achieves significant performance improvements on both the ST-VQA and TextVQA datasets and provides a novel paradigm for multimodal adversarial training.
\end{abstract}
\begin{keywords}
Scene-Text Visual Question Answering, multimodal reasoning, adversarial training, scene understanding
\end{keywords}

\section{Introduction}
\label{sec:intro}
The semantic information conveyed by scene text plays an indispensable role in performing real-world question-answering tasks. Building upon the success of Visual Question Answering (VQA)~\cite{antol2015vqa} research, the task of Scene-Text Visual Question Answering (ST-VQA)~\cite{singh2019towards,yang2021tap} has been introduced to answering questions related to scene text. In order to conduct joint reasoning of various semantic information, encompassing language, vision, and layout~\cite{xu2020layoutlmv2,biten2022latr}, a plethora of Transformer-based~\cite{vaswani2017attention} network architectures have been introduced to fuse information from different modalities. These architectures~\cite{yang2021tap,biten2022latr,li2022two,jin2022token,zeng2021beyond} enhance representational capabilities for ST-VQA tasks significantly through alignment of linguistic spatial features. 

Due to the fact that the text in scene images is recognized by Optical Character Recognition (OCR) systems~\cite{baek2019wrong},  most works in this domain~\cite{yang2021tap,biten2022latr,li2022two,jin2022token,zeng2021beyond} integrate OCR text with multimodal transformers. These methods typically involve large-scale pretraining followed by fine-tuning to adapt the model for question-answering tasks in text-rich scene images, often ignoring the inevitable OCR text recognition challenges. In practice, scene images may exhibit phenomena such as blurring, distortion, skewness, or uneven lighting, leading to erroneous character recognition by OCR systems, especially in cases of low-quality handwriting. Even when OCR systems correctly identify characters, discrete and semantically irrelevant OCR recognition results may impact the comprehension of the text semantics. Learning contextual relationships in erroneous and semantically unrelated OCR text will inevitably introduce harmful noise. Additionally, spatial location information in the ST-VQA task is often limited. Extensive and aggressive fine-tuning may result in overfitting of the model~\cite{gan2020large}.

\begin{figure*}[!t]
\vspace{-11pt}
\begin{center}
  \centerline{\includegraphics[width=18cm]{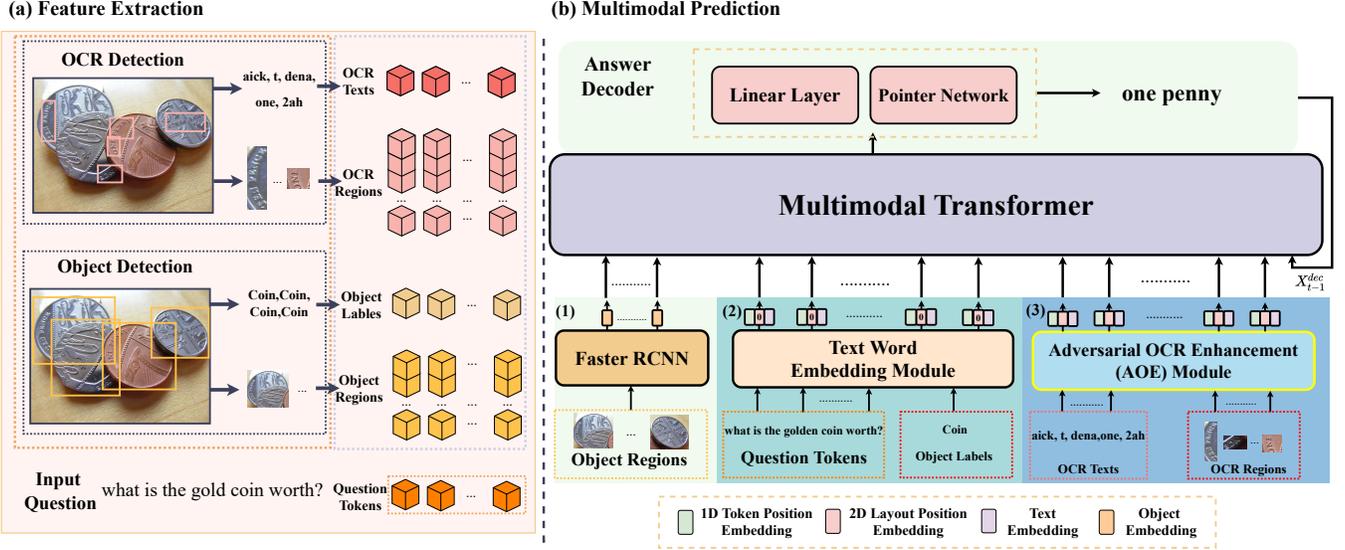}}
\end{center}
\vspace{-0.25in}
	\caption{
	An overview of our \textbf{ATS}. \textbf{(a)} We extract question tokens, visual objects, OCR tokens and OCR regions from an image and a question. \textbf{(b)} We perform adversarial training on the Multimodal Transformer to predict answers through iterative decoding.}
\label{fig:Proce}
\vspace{-0.15in}
\end{figure*}

In this paper, we focus on learning more generalized spatial semantic information in the embedding space of OCR modality and propose \textbf{ATS}: an \textbf{A}dversarial \textbf{T}raining framework for the \textbf{S}T-VQA task. Specifically, we introduce an Adversarial OCR Enhancement (AOE) module, employing adversarial perturbations at the multimodal embedding level for adversarial training. To mitigate semantic biases and potential errors in OCR texts, we enhance OCR tokens by modifying specific characters in the OCR system's recognition results and constructing fault-tolerant representations for OCR texts. Furthermore, we incorporate the Spatial-Aware Self-Attention (SASA) mechanism into the AOE module. SASA utilizes relative position representations of token pairs as two-dimensional position embeddings, providing a broader perspective for spatial modeling in OCR texts. Simultaneously, adversarial training enhances the model's generalization ability, thus effectively alleviating the overfitting problem. According to the results, ATS significantly outperforms existing methods with much less training data. In summary, the main contributions of our work are:
\begin{itemize}
\item We are the first to design adversarial training architectures for ST-VQA, which can effectively force the model to learn a fault-tolerant representation of OCR texts to improve the model's generalization performance.
\item Instead of OCR absolute position embedding, we utilize OCR relative position embedding to provide a broader perspective for spatial modeling of extended text contexts.
\item Our approach reaches a new state-of-the-art in extensive experimental tests on the popular ST-VQA and TextVQA benchmarks.
\end{itemize}

\section{PROPOSED METHOD}
As shown in Fig. \ref{fig:Proce}, the entire ATS architecture comprises feature extraction and multimodal prediction. In this section, we will sequentially introduce the embeddings of different modality features (Sec. \ref{feature}), the AOE module used for OCR enhancement (Sec. \ref{AOE}), and the process of multimodal adversarial training (Sec. \ref{adv}).

\subsection{Multimodal Features Embeddings} \label{feature}
\textbf{Question Feature Embedding.} 
Question Feature Embedding combines question word embedding and positional embedding, where positional embedding includes 1D Token Positional Embeddings and 2D Layout Positional Embeddings. The architecture of the Question Embedding Module is based on LayoutLMv2~\cite{xu2020layoutlmv2}. We believe that the question text is semantically linked to the text in the scene image, and the use of relative positional embedding in 2D space helps to align the question text with the scene image-text features.

Specifically, for a question of a given length $L$, we take each tokenized question token as a sequence. The final question feature embedding is the sum of the three embeddings. We can get the $i$-th($0 \leq i<L$) question feature inputs $x^{Q}_i$:\begin{equation}
x^{Q}_i = Q^w_i + Q^{1D}_i + Q^{2D}_i
\end{equation}
word embeddings $Q^w$ denotes the question word embeddings, $Q^{1D}$ represents the index of all tokens, and $Q^{2D}$ denotes the 2D bounding box coordinates of all tokens, all of which are set to 0 in Question Feature Embedding and Visual Feature Embedding. 

\textbf{Visual Feature Embedding.}
It has been empirically demonstrated in previous studies~\cite{li2022two},~\cite{jin2022token},~\cite{lu2021localize} that visual object features are not an essential factor for the ST-VQA task. To obtain the Object region features, we use the same Faster RCNN as previous methods: 
\begin{equation}
x_{i}^{obj} = LN(W_1obj^{fr}_i)+LN(W_2obj^{bx}_i)+obj^{lb}_i
\end{equation}
where $LN$ denotes layer normalization, $W$ represents the learned projection matrix, $obj^{fr}_i$ stands for the appearance feature of the object, $obj^{bx}_i$ corresponds to the bounding box feature, and $obj^{lb}_i$ is the LayoutLMv2 word embedding corresponding to the label of the object.

\textbf{OCR Feature Embedding.}
Similar to Question Feature Embedding, we categorize the extracted OCR features into textual, visual, and layout information as inputs to LayoutLMv2.

Formally, the $i$-th OCR token embedding input can be represented as $ocr^{txt}_i$, and its corresponding 2D spatial position can be denoted as:
\begin{equation}
ocr^{bx}_i=(\frac{x_i^{0}}{w},\frac{y_i^{0}}{h},\frac{x_i^{1}}{w},\frac{y_i^{1}}{h},h,w)
\end{equation}
where $(x_i^{0},y_i^{0})$ are the coordinates of the upper-left corner of the 2D spatial position corresponding to the $i$-th token, $(x_i^{1},y_i^{1})$ are the coordinates of the lower-right corner, and $(h, w)$ correspond to the height and width. A linear projection layer is applied to visual feature embeddings, making them share the same embedding layer with text embeddings. Since the standalone Faster RCNN cannot capture positional information, we also add the 1D Position Embeddings of the visual embeddings to the 1D Position Embeddings of the text token embeddings. The 2D Layout Position Embeddings will be set to 2D bounding box coordinates of OCR tokens and visual objects. The final OCR feature is represented as:
\begin{align}
    x_{i}^{ocr} = ocr^{1D}_{i-1}+ocr^{2D}_{i}+ocr^{txt/vis}_i
\end{align}
where $ocr^{1D}$ represents the 1D Token Positional Embeddings, $ocr^{2D}$ corresponds to the 2D Layout Positional Embeddings, and $ocr^{txt/vis}$ denotes the Text/Faster RCNN Token Embeddings.

After embedding the Question, Object and OCR features, the decoded output from the previous step is also embedded and fed into the Transformer. The final input embedding is:
\begin{equation}
input = [X^Q, X^{obj}, X^{ocr}, X^{dec}_{t-1}]
\end{equation}
where $X^Q$ represents the Question Feature Embedding, $X^{obj}$ is the Visual Feature Embedding, $X^{ocr}$ denotes the OCR Feature Embedding, and $X^{dec}_{t-1}$ represents the decoding output from the previous step.

\begin{figure}[t]
    \centering
    \includegraphics[width=8.5cm]{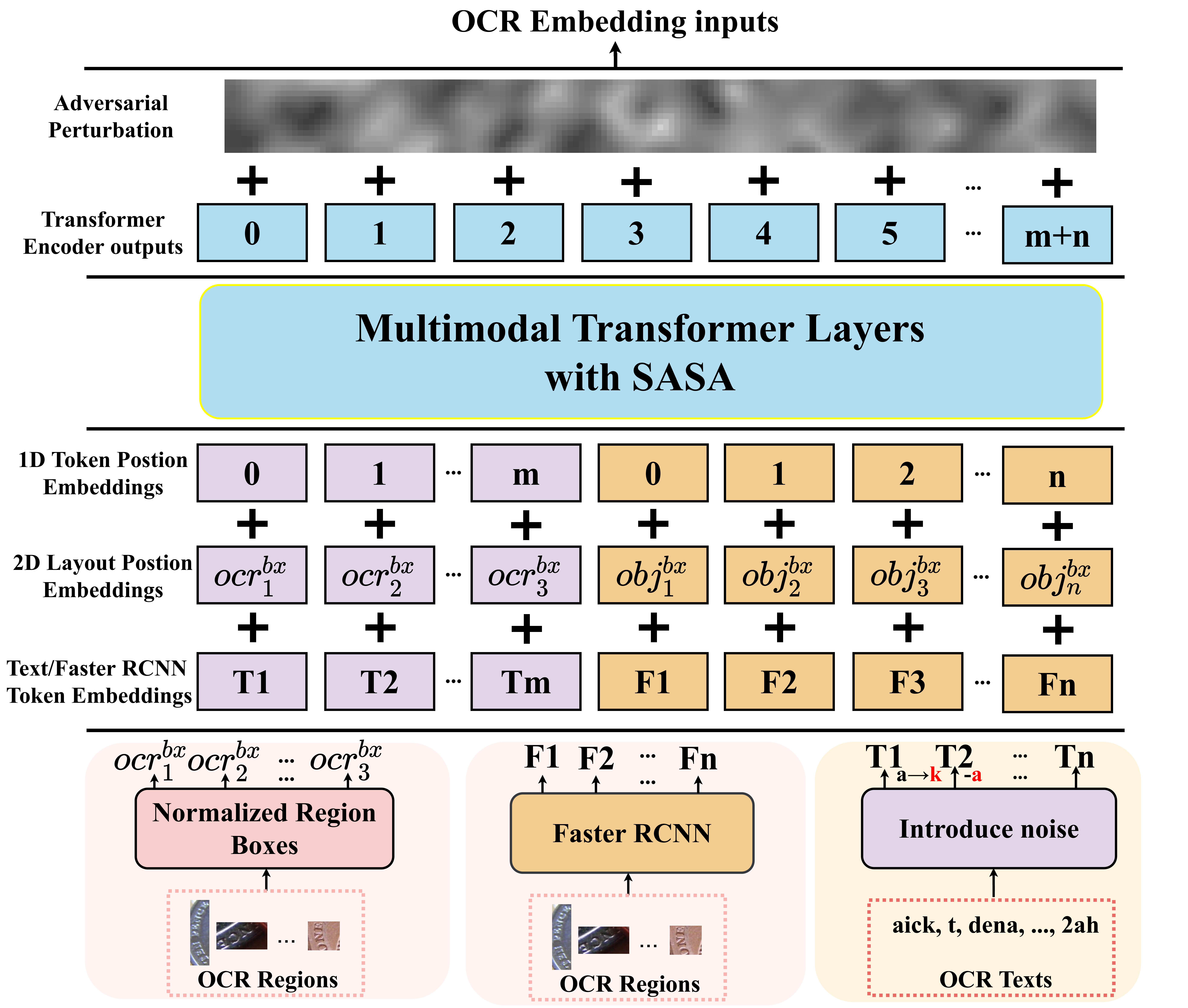}
    \caption{The architecture of Adversarial OCR Enhancement (AOE) Module.}
    \label{fig:Aoe}
    \vspace{-0.15in}
\end{figure}

\subsection{Adversarial OCR Enhancement Module} \label{AOE}
As shown in Fig. \ref{fig:Aoe}, in order to enhance the model's tolerance to potential OCR errors, we propose the Adversarial OCR Enhancement (AOE) module to strengthen the model's robust representation of the input OCR texts.

For the OCR Region, we extract the coordinates of bounding boxes and visual features from the Faster RCNN. For the OCR Texts, following the architecture of ~\cite{jin2022token}, to maintain semantic consistency, we set a threshold $\lambdav$ to control OCR Token variations. Assuming that the changed OCR Token can be found in the CharBERT~\cite{ma2020charbert} dictionary, we will introduce character noise to OCR Tokens by deleting, adding, or replacing specific tokens with similar words from the dictionary. If the token is not in the dictionary, we search for the word with the shortest edit distance in the dictionary for replacement. 

In the AOE Transformer blocks, We adopt the Spatial-Aware Self-Attention (SASA) mechanism to replace the original self-attention mechanism in the multimodal Transformer. For a single head in a single self-attention layer with hidden size $d_k$, the original self-attention mechanism computes the correlation between two vectors by projecting two input vectors to find the self-attention score:
\begin{equation}
    a_{i,j} = \frac{(Q^{ocr}_{i}W^Q) \cdot (K^{ocr}_{j}W^k)^\mathrm{T}}{\sqrt{d_k}}
\end{equation}
where $W^Q$, $W^K$, $W^V$ are projection matrixs, $Q^{ocr}_{i}$ is the query vector corresponding to OCR token $i$ and $K^{ocr}_{j}$ is the key vector of OCR token $j$. 
We model semantic relative positions and spatial relative positions as bias terms and combine their positional biases with the attention scores computed by the original self-attention mechanism to obtain spatial-aware attention scores:
\begin{equation}
    \widehat{a}_{i,j} = a_{i,j} + \widehat{a}^{1D}_{j-i} + \widehat{a}^{2D_x}_{x_j-x_i} + \widehat{a}^{2D_y}_{y_j-y_i}
\end{equation}
where $(x_{i},y_{i}), (x_{j},y_{j})$ denote the $i$-th and the $j$-th OCR token bounding box, $\widehat{a}^{1D}_{j-i}$, $\widehat{a}^{2D_x}_{x_j-x_i}$ and $\widehat{a}^{2D_y}_{y_j-y_i}$ are the 1D and 2D relative position biases. The spatial-aware attention score replaces the Transformer self-attention score in the AOE module.

The output vectors are represented as weighted averages of the projected value vectors, where the weights are their spatial perceptual attention scores relative to the normalized one:

\begin{equation}
    b_i = \sum_{j}{\frac{\exp(\widehat{a}_{i,j})}{\sum_{k}{\exp(\widehat{a}_{i,k})}}} \cdot K^{ocr}_{j} \cdot W^V
\end{equation}


Given OCR textual feature embedding $\xv^{txt}$, visual feature embedding $\xv^{vis}$ and bounding box embedding $\xv^{bbox}$, we can get the OCR feature embedding $\xv^{ocr}{'} = (\xv^{txt} + \deltav, \xv^{vis}, \xv^{bbox})$ after adding noise $\deltav$.
The OCR Transformer outputs result can be expressed as:
\begin{equation}
    \zv^{ocr} = f^{ocr}_{\thetav}(\xv^{txt} + \deltav \cdot (1-\lambdav)^{k} \cdot \lambdav^{1-k}, \xv^{vis}, \xv^{bbox})
\end{equation}
where $f^{ocr}$ is the Transformer outputs, and $\theta$ is all learnable parameters in the embedding matrix. The probability of modifying OCR tokens obeys the Bernoulli distribution. Subsequently, we add the adversarial perturbation $\deltav_{ocr}$ in the embedding space of the OCR modality. It is worth noting that the adversarial perturbation will be introduced at the output of the AOE transformer encoder to ensure that other feature information is not affected during adversarial training. The OCR embedding inputs will become $\xv^{ocr}{''} = \zv^{ocr}+\deltav_{ocr}$.

\setlength{\textfloatsep}{10cm}
\begin{algorithm*}[t!]
	\caption{
		Multimodal Adversarial Training used in \textsc{ATS}.
	}
	\label{alg:freemat}
	\begin{algorithmic}[1]
		\Require Training sample subsets $\dv_{t}=\{(\xv^{txt}, \xv^{obj}, \xv^{ocr}{''}, \xv^{dec}_{t-1},\yv_{gt})\}$, learning rate $\tau$, ascent steps $K$, ascent step size $\alpha$, perturbation bound $\lambda$,  
		\State Initialize $\thetav$
		\For{epoch $= 1 \ldots N_{ep}$}
		\For{minibatch $b\subseteq N$}
        \State $\gv_0 \gets 0$,\,\, $\deltav_0 \gets \frac{1}{\sqrt{N_{\deltav}}} U(-\lambda,\lambda)$
		\For{$t =1 \ldots K$}
		\State Accumulate gradient of parameters $\thetav$ given $\deltav_{ocr,t-1}$
		\State \qquad  $\gv_t \gets \gv_{t-1} + \frac{1}{K}\E_{(\xv^{Q}, \xv^{obj}, \xv^{ocr}{''}, \xv^{dec}_{t-1}, \yv_{gt})} \sim d[\nabla_{\theta} \L_{pred}(\theta) + \L_{kl}(\theta)] \text{ and } d \subseteq b $
		\State Update the perturbation $\deltav_{ocr}$ via gradient ascend
		\State \qquad  $\yv_{pred} = f_{\thetav}(\xv^{Q}, \xv^{obj}, \zv^{ocr}, \xv^{dec}_{t-1})$
        \State \qquad  $\gv_{ocr} \gets  \nabla_{\deltav_{ocr}} \, [ L_{pred}(f_{\thetav}(\xv^{Q}, \xv^{obj}, \zv^{ocr}+\deltav_{ocr}, \xv^{dec}_{t-1}), \yv_{gt}) + L_{kl}(f_{\thetav}(\xv^{Q}, \xv^{obj}, \zv^{ocr}+\deltav_{ocr}, \xv^{dec}_{t-1}),\yv_{pred}) ]$
		\State \qquad $\deltav_{ocr,t} \gets \Pi_{\lVert \deltav_{ocr} \rVert_F\le \lambda}(\deltav_{ocr,t-1} + \alpha \cdot \gv_{ocr} / \lVert \gv_{ocr} \rVert_F$)
		\EndFor
        \State  $\thetav \gets \thetav - \tau \gv_K$
		\EndFor
		\EndFor
	\end{algorithmic}
\end{algorithm*}
\vspace{0.1cm}

\subsection{Multimodal Adversarial Training}\label{adv}
In ATS, we use adversarial training as an effective regularization to improve model generalization. Specifically, the goal of our adversarial training is to minimize the following objective: 
\begin{equation}
    \min_{\thetav} \E_{(\xv^{Q}, \xv^{obj}, \xv^{ocr}{''}, \xv^{dec}_{t-1}, \yv_{gt})} \sim \Dcal [\Lcal_{pred}(\thetav) + \Lcal_{kl}(\thetav)]
\end{equation}
Following previous works~\cite{yang2021tap,jin2022token}, given the set of input subsets $\dv_{t} = [\xv^Q, \xv^{obj}, \xv^{ocr}{''}, \xv^{dec}_{t-1}]$, model prediction $\yv_{pred}$ and prediction with the addition of adversarial perturbations $\yv_{pred}{'}$ can be denoted as $f_{\thetav}(\xv^{Q}, \xv^{obj}, \zv^{ocr})$ and $f_{\thetav}(\xv^{Q}, \xv^{obj}, \xv^{ocr}{''})$, respectively. $\Lcal_{pred}(\theta)$ is set to be the binary cross-entropy loss for multi-label classification for predicting answers, which is defined as follows:
\begin{equation}
    \pv = \frac{1}{1+\exp(-f_{\thetav}(\xv^Q, \xv^{obj}, \xv^{ocr}{''}, \xv^{dec}_{t-1}))}
\end{equation}
\begin{equation}
    \Lcal_{pred}(\thetav) = -(\yv_{gt}\log(\pv))+(1-\yv_{gt})\log(1-\pv)))
\end{equation}
$\Lcal_{kl}(\theta)$ denotes fine-grained adversarial regularization, $\yv_{gt}$ is the ground-truth answer. Taking the adversarial perturbation $\delta_{ocr}$ for OCR feature as an example:
\begin{align}
    \Lcal_{kl}(\thetav) &= \max_{||\deltav_{ocr}|| \leq \lambda} KL(\yv_{pred}{'},\yv_{pred})
\end{align}
where $KL(\yv',\yv) = kl(\yv'||\yv)+kl(\yv||\yv')$, $kl$ denotes the Kullback-Leibler Divergence.

As shown in Algorithm \ref{alg:freemat}, we employ Projected Gradient Descent (PGD) to update model parameters during training. To mitigate the significant computational cost associated with calculating gradients and updating input samples at each iteration, we overlook the intricacies and interpretability of generating adversarial samples and focus only on the impact of adversarial training on the final model prediction results. In each iteration, $K$ separate instances of PGD are individually configured to craft adversarial embeddings. Each instance accumulates the corresponding parameter gradient: $\frac{1}{K}\E_{(\xv^{Q}, \xv^{obj}, \xv^{ocr}{''}, \xv^{dec}_{t-1}, \yv_{gt})} \sim d[\nabla_{\theta} \L_{pred}(\theta) + \L_{kl}(\theta)]$, and after each iteration, we accumulate all PGD parameter gradients $\nabla_{\theta} L$ and update the model parameters $\theta$ based on the cumulative gradient.

\begin{table}[!t]
\setlength{\abovecaptionskip}{0cm}
\setlength{\belowcaptionskip}{-0.4cm}
\normalsize
\begin{center}
\footnotesize
\setlength{\tabcolsep}{1.6mm}
\caption{Results on the ST-VQA Dataset. '$\dagger$$\dagger$' refers to a larger model with 8 layers of LayoutLMv2 and 12 layers of Transformer.}
\bgroup
\def\arraystretch{1.1}
        \begin{tabular}{lcccc}
        \toprule
        \textbf{Method} & \textbf{Extra Data} & \textbf{Val Acc.(\%)} & \textbf{Val ANLS} & \textbf{Test ANLS}\\
        \midrule
        M4C~\cite{hu2020iterative} & \ding{55} & 38.05 & 0.472 & 0.462 \\
        BOV~\cite{zeng2021beyond} & \ding{55} & 40.18 & 0.500 & 0.472 \\
        SA-M4C~\cite{kant2020spatially} & \ding{55} & 42.23 & 0.512 & 0.504 \\
        TAP~\cite{yang2021tap} & \ding{55} & 45.29 & 0.551 & 0.543 \\
        LOGOS~\cite{lu2021localize} & \ding{55} & 48.63 & 0.581 & 0.579 \\
        TWA~\cite{jin2022token} & \ding{55} & 48.07 & 0.563 & 0.577 \\
        TWF~\cite{li2022two} & \ding{55} & 50.49 & 0.598 & 0.587 \\
        \rowcolor{LightGray} ATS & \ding{55} & 52.56 & 0.617 & 0.614 \\
        \rowcolor{LightGray} ATS & TextVQA & 53.24 & 0.636 & 0.631 \\
        \rowcolor{LightGray} ATS$^\dagger$$^\dagger$ & TextVQA & \textbf{54.84} & \textbf{0.642} & \textbf{0.638} \\ 
        \bottomrule
    \end{tabular}

\egroup
\tiny
\label{table:stvqa}
\vspace{-10cm}
\end{center}
\end{table}

\begin{table*}[!t]
\normalsize
\setlength{\abovecaptionskip}{0cm}
\setlength{\belowcaptionskip}{-0.4cm}
\begin{center}
\footnotesize
\setlength{\tabcolsep}{3.6mm}
\caption{Comparison of results with state-of-the-art methods on the TextVQA dataset. The top of the table shows the results using Rosetta OCR and training with only the TextVQA dataset. The bottom of the table shows the results with unlimited settings.}
\bgroup
\def\arraystretch{1.1}
        \begin{tabular}{llcccccc}
        \toprule
        \textbf{Method} & \textbf{OCR System} & \textbf{Extra Data} & \textbf{Data Size} & \textbf{Param. Size} & \textbf{Val Acc.(\%)} & \textbf{Test Acc.(\%)}\\
        \midrule
        M4C~\cite{hu2020iterative} & Rosetta-en & \ding{55} & 0.03M & 150M & 39.40 & 39.01 \\
        TAP~\cite{yang2021tap} & Rosetta-en & \ding{55} & 0.03M & 160M & \eat{\gree{44.06}}44.06 & - \\
        LaTr~\cite{biten2022latr}  & Rosetta-en & \ding{55} & 0.03M & 311M & 44.06  & - \\
        BOV~\cite{zeng2021beyond} & Rosetta-en & 
        \ding{55} & 0.03M & - & 40.90  & 41.23 \\
        \rowcolor{LightGray} ATS & Rosetta-en & \ding{55} & 0.03M & 198M & \textbf{47.46} & \textbf{-} \\
        \midrule
        SA-M4C~\cite{kant2020spatially} & Google-OCR & ST-VQA & 0.06M & 150M & 45.4 & 44.6 \\
        BOV~\cite{zeng2021beyond} &  SBD-Trans OCR  & ST-VQA & 0.06M &  - & - & 45.51 \\
        TAP~\cite{yang2021tap} & Microsoft-OCR & \ding{55} & 0.03M & 160M & 49.91 & 49.71 \\
        TAP~\cite{yang2021tap} & Microsoft-OCR & ST-VQA, TextCaps, OCR-CC & 1.5M & 160M & 54.71 & 53.97 \\
        LOGOS~\cite{lu2021localize} & Microsoft-OCR & ST-VQA &  0.06M & -  & 51.53 & 51.08 \\
        TWF~\cite{li2022two} & Microsoft-OCR & ST-VQA & 0.06M & - & 54.33 & 54.47 \\
        LaTr~\cite{biten2022latr} & Amazon-OCR & \ding{55} & 0.03M & 311M & 52.29 & - \\
        \rowcolor{LightGray} ATS & Microsoft-OCR & \ding{55} & 0.03M & 198M & 55.72 & 55.66 \\
        \rowcolor{LightGray} ATS & Microsoft-OCR & ST-VQA & 
        0.06M & 198M & 56.89 & 56.37 \\
        \rowcolor{LightGray} ATS$^\dagger$$^\dagger$ & Microsoft-OCR & ST-VQA & 0.06M & 266M & \textbf{57.68} & \textbf{56.99} \\
        \bottomrule
    \end{tabular}
\egroup
\tiny
\label{table:textvqa}
\vspace{-0.35cm}
\end{center}
\end{table*}

\begin{table}[!t]
\setlength{\abovecaptionskip}{0cm}
\setlength{\belowcaptionskip}{0cm}
\normalsize
\begin{center}
\small
\footnotesize
\setlength{\tabcolsep}{1.5mm}
\caption{Ablation Studies on the TextVQA dataset. Our baseline is TAP (49.91\%). ATS$^\dagger$  represents ATS trained with additional ST-VQA dataset.}
\bgroup
\def\arraystretch{1.1}
        \begin{tabular}{cccccc}
        \toprule
        \textbf{Method} &\textbf{Token-Noise} &\textbf{2D Layout} &\textbf{SASA} &\textbf{ADV OCR} &\textbf{Acc.(\%)} \\ 
        \midrule
        \multirow{7}{*}{ATS} &\ding{55} &\ding{55} &\ding{55} & \ding{55} & 49.91 \\
        &\ding{55} & \ding{55}  & \ding{55} & \checkmark & 50.33 \\
         &\ding{55} & \checkmark & \ding{55} & \ding{55} & 50.54 \\
         &\checkmark & \ding{55} & \ding{55} & \ding{55} & 51.72 \\
         &\ding{55} &\checkmark  & \checkmark & \ding{55} & 52.54\\
        
        &\checkmark & \checkmark  & \ding{55} & \ding{55} & 53.82 \\

         &\checkmark &\checkmark  & \checkmark & \ding{55} & 54.68 \\
         &\checkmark & \checkmark  & \checkmark & \checkmark & \textbf{55.72} \\
         \midrule
        \multirow{3}{*}{ATS$^\dagger$} &\ding{55} &\ding{55} &\ding{55} & \ding{55} & 50.57 \\
        &\checkmark &\checkmark &\checkmark & \ding{55} & 55.08 \\
        &\checkmark & \checkmark  & \checkmark & \checkmark & \textbf{56.89} \\
        \bottomrule
    \end{tabular}
\egroup
\tiny
\label{tab:ablation1}
\vspace{-10cm}
\end{center}
\end{table}

\section{Experiments} \label{fuc}
\subsection{Experiment Settings}
\textbf{Datasets and Metrics.} We conducted experiments on the ST-VQA and TextVQA datasets, maintaining the same settings as in previous works~\cite{yang2021tap,jin2022token}. We use the accuracy of soft voting among 10 answers as the evaluation criterion. The evaluation standard for the ST-VQA dataset includes the addition of the Average Normalized Levenshtein Similarity (ANLS) metric. ANLS is defined as $1 - d_L(\yv_{pred},\yv_{gt}) / max(|\yv_{pred}|,|\yv_{gt}|)$, where $\yv_{pred}$ is the prediction answer, $\yv_{gt}$ is the ground-truth answer, $d_l$ is the edit distance.

\textbf{Implementation details.} Our base model consists of a 6-layer LayoutLMv2 and a 4-layer 12-attention-head multimodal Transformer. The multimodal Transformer is initialized from the beginning. The weight of Kullback-Leibler Divergence for adversarial perturbation in the AOE module is empirically set to 1.5. We use the AdamW~\cite{loshchilov2017decoupled} optimizer for training, and the batch sizes for pre-training and fine-tuning are set to 48 and 8, respectively. All training processes are performed for 48,000 iterations, with the Transformer learning rate set to 1e-4, the Warm-Up Learning Rate to 0.2, and the number of Warm-Up iterations to 1000.

\subsection{Experiment Results}

\textbf{ST-VQA Results.} The performance of our approach on ST-VQA can be seen in Table \ref{table:stvqa}. The performance of our approach on the validation and test sets shows a significant improvement over the baseline TAP~\cite{yang2021tap} (from 45.29\% and 0.551 to 52.56\% and 0.617, respectively) without using extra data. When trained on a larger model, our approach achieves a final accuracy and ANLS score of 54.84\% and 0.642, which outperforms TWF-Base (50.49\%, 0.598) by 4.35\% and 0.044, respectively, validating the stability and effectiveness of the method.

\textbf{TextVQA Results.} The performance of our approach on TextVQA can be seen in Table \ref{table:textvqa}. In the case of using less training data and parameters, our approach achieves an accuracy of 55.72\% on the validation set, surpassing the previous state-of-the-art method~\cite{li2022two} by 2.39\%. With the help of the extra ST-VQA dataset, our large model achieves a new best accuracy of 57.68\% on the validation set.

\begin{figure*}[t]
    \centering
    \includegraphics[width=17.7cm]{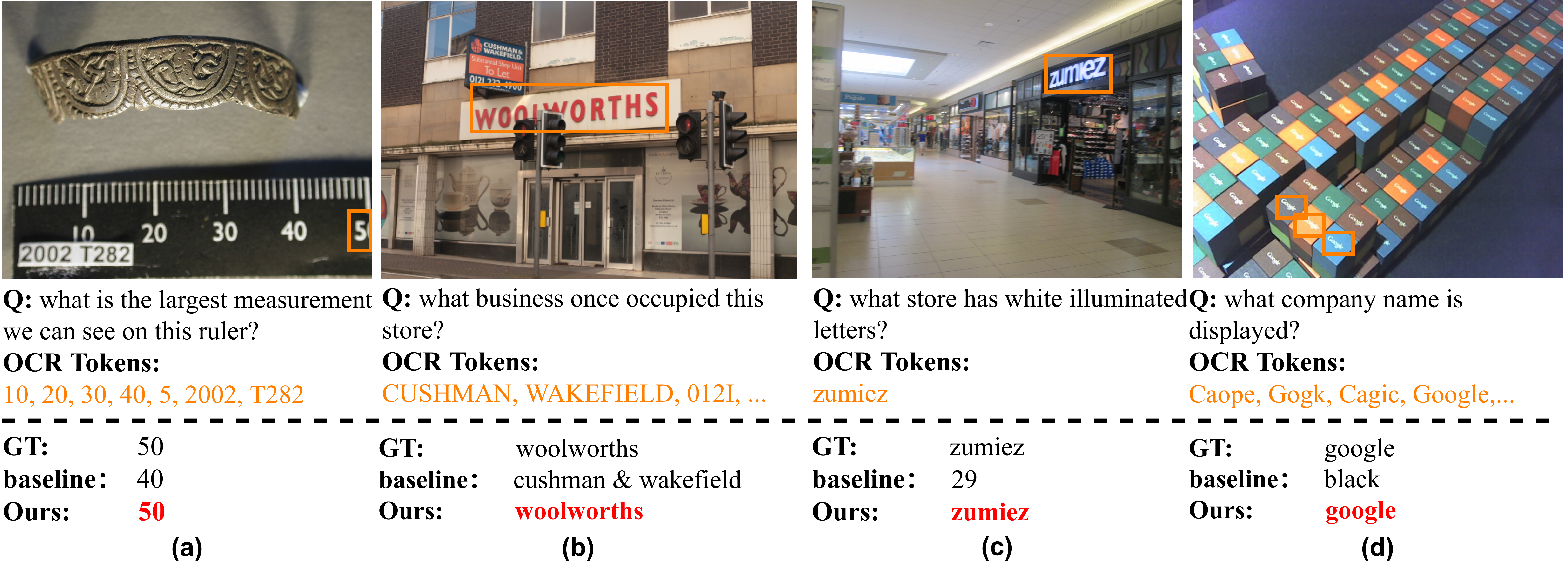}
    \caption{Qualitative results on TextVQA. For better visualization, the box in each image shows the most relevant text to the question. Best viewed in zoom.}
    \label{fig:qualit}
    \vspace{-0.15in}
\end{figure*}

\textbf{Qualitative Evaluation.} In Fig \ref{fig:qualit}, we present some qualitative samples from the TextVQA validation set, demonstrating the contrast between our method and the baseline in cases of OCR recognition errors and spatial location irrelevance. As shown in Fig \ref{fig:qualit}(a), (b), missing and occluded text in the scene images leads the model to extract incorrect semantic information. Employing adversarial training through OCR noise introduction, our approach compels the model to learn tolerant representations of OCR tokens, thus inferring the correct answers. Fig \ref{fig:qualit}(c), (d) illustrates the impact of our 2D Layout and SASA. Our model considers the semantic and spatial modeling relationship of the text and OCR tokens to provide the correct answer.


\subsection{Ablation Studies}
We perform a number of ablation studies on TextVQA dataset to examine the effectiveness of each component of ATS and ATS$^\dagger$ as shown in Table \ref{tab:ablation1}. Token-Noise, 2D Layout, SASA, and ADV OCR respectively denote introducing character noise, 2D layout information, Spatial-Aware Self-Attention mechanism, and introducing OCR adversarial perturbation. As can be seen, the completion results become worse in all the four ablation cases, indicating the effectiveness of the corresponding components.

\section{Conclusion}
In this work, we propose an adversarial training architecture tailored for ST-VQA tasks. Our approach utilizes an Adversarial OCR Enhancement (AOE) module for adversarial training to enhance the fault-tolerant representation of OCR texts. Additionally, to alleviate overfitting arising from insufficient spatial positional information in OCR labels, we introduce a Spatial-Aware Self-Attention (SASA) mechanism into the AOE module, designed to assist the model in better capturing spatial relationships among OCR tokens. Extensive experiments on two benchmarks validate the effectiveness and superiority of our method.

\section{Acknowledgement}
The authors express their gratitude to the National Natural Science Foundation of China (62306247, 62376231), China Postdoctoral Science Foundation (2022M722630), the Key R\&D Project of Sichuan Province (2022YFG0028) and Central Universities Basic Research Operating Costs Project (2682023CX034).

\bibliographystyle{IEEEbib}
\footnotesize
\bibliography{ref.bib}

\end{document}